\documentclass[10pt, conference, letterpaper]{IEEEtran}

\IEEEoverridecommandlockouts
\usepackage{enumitem}
\usepackage{cite}

\usepackage{pgfplots}
\pgfplotsset{compat=1.18}
\usepackage{amsmath,amssymb,amsfonts}
\usepackage{amsthm}
\usepackage{algorithm}
\usepackage{algpseudocode}
\usepackage{svg}
\usepackage{graphicx}
\usepackage{textcomp}
\usepackage{subfig}
\usepackage{multirow}
\usepackage{geometry}
\usepackage{tikz}
\usetikzlibrary{arrows.meta,positioning,fit,calc,shapes.geometric,shapes.symbols}

\newtheorem{proposition}{Proposition}

\newgeometry{top=1in, bottom=0.75in, left=0.75in, right=0.75in}

\def\BibTeX{{\rm B\kern-.05em{\sc i\kern-.025em b}\kern-.08em
    T\kern-.1667em\lower.7ex\hbox{E}\kern-.125emX}}

\AtBeginDocument{%
  \restoregeometry 
  \newgeometry{top=0.75in, bottom=0.75in, left=0.75in, right=0.75in} 
}

\usepackage{colortbl}
\usepackage{soul}
\usepackage[table,xcdraw]{xcolor}
\usepackage{float}
\usepackage[colorlinks=true, allcolors=blue]{hyperref}%
\usepackage{booktabs}

\begin{document}



\title{\LARGE \bf
SERN: Simulation-Enhanced Realistic Navigation for Multi-Agent Robotic Systems in Contested Environments}

\title{\LARGE \bf
SERN: Bandwidth-Adaptive Cross-Reality Synchronization for Simulation-Enhanced Robot Navigation}

\author{Jumman Hossain$^{1}$, Emon Dey$^{1}$, Snehalraj Chugh$^{1}$, Masud Ahmed$^{1}$, MS Anwar$^{1}$, Abu-Zaher Faridee$^{1,2}$, \\ Jason Hoppes$^{3}$, Theron Trout$^{3}$, Anjon Basak$^{3}$, Rafidh Chowdhury$^{3}$, Rishabh Mistry$^{3}$, Hyun Kim$^{3}$,  Jade \\Freeman$^{4}$, Niranjan Suri$^{4}$, Adrienne Raglin$^{4}$, Carl Busart$^{4}$, Anuradha Ravi$^{1}$, and Nirmalya Roy$^{1}$%
\thanks{*This work has been supported by U.S. Army Grant \texttt{\#W911NF2120076} and NSF CNS EAGER Grant \texttt{\#2233879}.}%
\thanks{$^{1}$\scriptsize Department of Information Systems, University of Maryland, Baltimore County, USA. {\tt\scriptsize \{jumman.hossain,edey1,schugh1,mahmed10,nroy\}@umbc.edu}.}%
\thanks{$^{2}$Amazon Inc., USA. {\tt\scriptsize abufari@amazon.com}.}%
\thanks{$^{3}$Stormfish Scientific Corporation, {\tt\scriptsize theron.trout@stormfish-sci.com}.}%
\thanks{$^{4}$DEVCOM Army Research Lab, USA. {\tt\scriptsize \{jade.l.freeman2.civ, timothy.c.gregory6.civ\}@army.mil}.}%
}
\maketitle

\begin{abstract}
Simulation and physical robots can complement each other during field operation, but this only works when the virtual side stays well aligned with the real system under delayed, or limited communication. We present \textbf{SERN} (\underline{S}imulation \underline{E}nhanced \underline{R}ealistic \underline{N}avigation), a framework for bandwidth adaptive cross reality synchronization in simulation enhanced robot navigation. SERN tightly couples a high fidelity virtual twin with a physical robot through three main components: (i) a virtual twin initialized from geospatial and sensor data and continuously corrected using live robot telemetry, (ii) a predictor corrector synchronization pipeline that combines model based propagation with adaptive PD correction, and (iii) a bandwidth adaptive SERN ROS Bridge that prioritizes critical topics under constrained links. We also introduce a multi metric cost function that balances latency, reliability, computation, and bandwidth when allocating communication resources. We show that when physical and virtual input mismatch remains bounded, synchronization error also remains bounded under moderate packet loss and latency. In experiments, SERN reduces end to end message latency by 15\% to 25\% and processing load by about 15\% compared with a standard ROS setup, while maintaining tight real virtual alignment with less than 5\,cm positional error and less than $2^\circ$ rotational error. On a navigation task, SERN achieves 95\% success, compared with 85\% for a real only setup and 70\% for a simulation only setup, with fewer interventions and shorter time to goal. 


\end{abstract}
\begin{IEEEkeywords}
digital twins,
cross reality integration,
real time synchronization,
ROS communication,
bandwidth adaptive networking,
constrained communication,
robot navigation,
cyber physical systems
\end{IEEEkeywords}
\section{Introduction}

\begin{figure}[!htb]
    \centering
    \includegraphics[width=\linewidth]{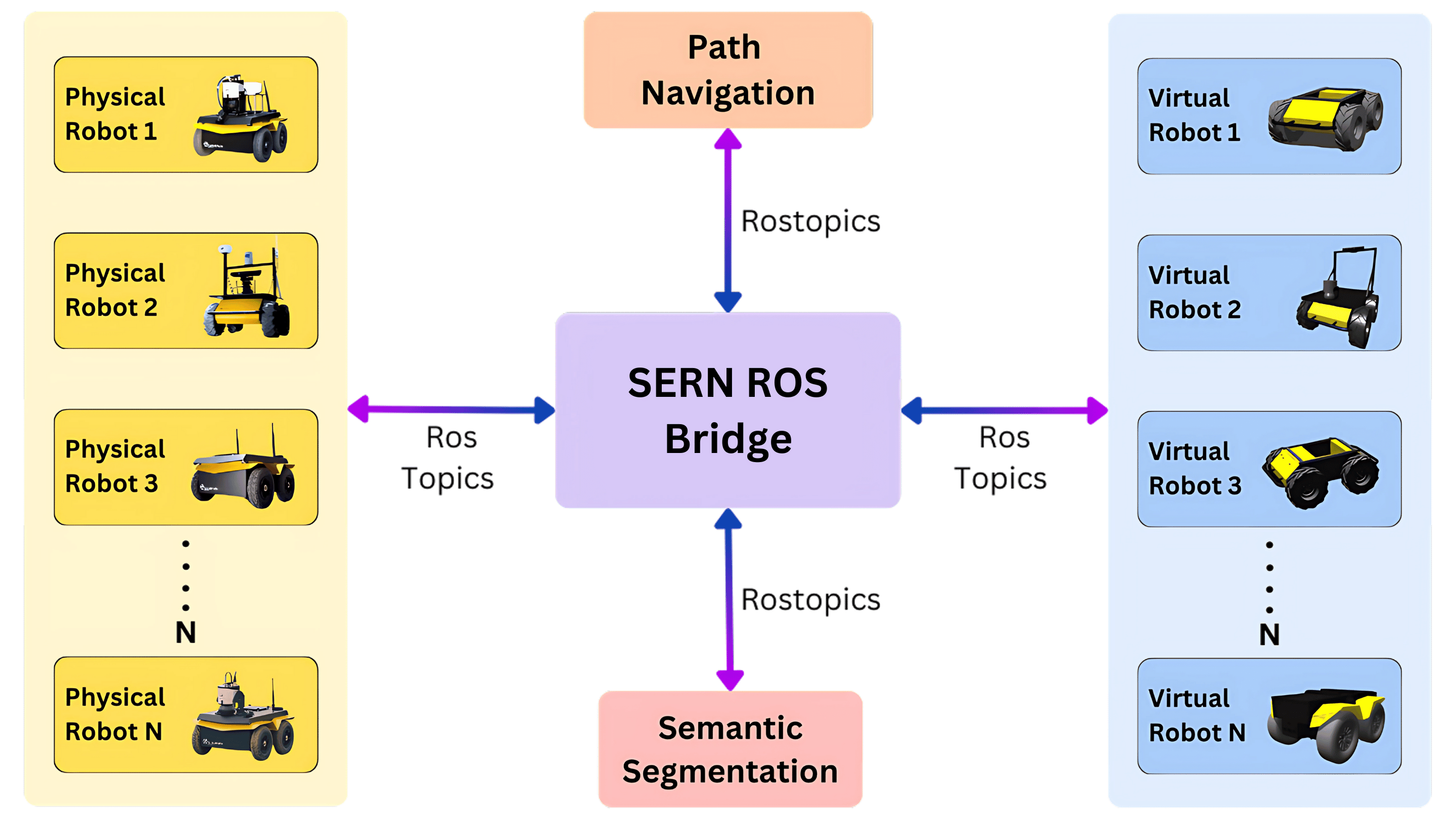}
    \caption{SERN enables bi-directional communication and synchronization between physical robots and a virtual twin, supporting coordinated robot operations across separate networks.}
    \label{fig:ros_physical_virtual_integration}
\end{figure}

Robots that rely on remote communication often have to operate with delayed, lossy, or limited links. In these settings, the main difficulty is not only how to run the robot safely, but also how to keep remote monitoring, planning, and decision support consistent with what the robot is actually seeing and doing. This becomes especially important when a simulation side twin is used to support navigation, since the twin is only useful when it stays close to the physical system over time~\cite{segovia2022digitaltwins,frasheri2023timediscrepancy,gil2026cosimdtrobotics}. A simulation side twin can still be valuable in these conditions. It can provide short horizon prediction, support remote visualization, and help maintain situational awareness when onboard sensing and remote computation need to work together~\cite{tahir2023collaborativetwin,bergs2025dtnavigation}. At the same time, a twin can fall out of sync when communication becomes unreliable. Delayed telemetry, packet loss, and temporary disconnections can all cause the virtual side to fall out of sync with the robot, which weakens the value of the simulation and can mislead downstream decisions.

In practice, the system must keep the physical and virtual states aligned while also using limited bandwidth carefully enough to preserve the most important information~\cite{frasheri2023timediscrepancy,gielis2022critical}. Existing approaches usually rely on either simulation or physical deployment alone, and each has clear limits. Simulation is flexible and useful for look ahead support, but it cannot fully capture field conditions. Physical testing provides grounded feedback, but it is expensive, slower to iterate, and harder to scale. A useful middle ground is to integrate a physical robot with a continuously updated virtual counterpart, so that the twin can support planning and monitoring while live sensing keeps the system grounded in the real world~\cite{mohaghegh2025sim2realdt,bergs2025dtnavigation}. The challenge is that this connection must remain reliable under constrained communication, where bandwidth, latency, and loss directly affect synchronization quality~\cite{gielis2022critical,coronado2024industry5ros}.

To address this problem, we present \textbf{SERN} (Simulation Enhanced Realistic Navigation), a framework for bandwidth adaptive cross reality synchronization in simulation enhanced robot navigation. SERN integrates a high fidelity virtual twin with a physical robot through a bi-directional communication layer and a synchronization loop that are designed to work together (Fig.~\ref{fig:ros_physical_virtual_integration}). The system builds on an AuroraXR based communication path~\cite{dennison2022auroraxr} and extends it with bandwidth aware topic scheduling, explicit support for synchronization critical state updates, and a predictor correction mechanism that helps the twin remain aligned when the connection degrades. Instead of handling communication and synchronization separately, SERN brings them together as tightly connected parts of one system.

\noindent \textbf{Our main contributions are}:
\begin{itemize}
    \item \textbf{Virtual twin with live correction.} A high fidelity virtual twin initialized from geospatial and sensor data and continuously updated with live robot telemetry, reducing manual setup and helping keep the virtual state close to the physical robot during operation.
    
    \item \textbf{Bandwidth adaptive SERN ROS Bridge.} An AuroraXR based bridge~\cite{dennison2022auroraxr} extended with priority aware forwarding for critical ROS topics under constrained links, together with buffering, replay, and runtime topic discovery to support changing stream availability without manual reconfiguration.
    
    \item \textbf{Predictor correction synchronization.} A synchronization pipeline that combines model based state propagation with adaptive PD correction to reduce mismatch between the physical robot and the virtual twin. We also provide a finite interval error bound for temporary communication gaps when the control mismatch remains bounded.
    
    \item \textbf{Multi metric runtime adaptation.} A cost function that balances latency, reliability, computation, and bandwidth when selecting bridge behavior at runtime, allowing the system to preserve synchronization critical traffic while reducing unnecessary communication load.
    
    
\end{itemize}

In our experiments, SERN reduces end-to-end message latency by 15\% to 25\% and processing load by about 15\% relative to a standard ROS setup, while maintaining tight real virtual alignment with less than 5\,cm positional error and less than $2^\circ$ rotational error under moderate delay and loss. In a navigation task, SERN achieves 95\% success, compared with 85\% for a real only setup and 70\% for a simulation only setup, with fewer interventions and shorter time to goal. These results show that a simulation enhanced cross reality framework can improve situational awareness and navigation robustness when communication is limited.
\section{Related work}
This section reviews the relevant prior work and highlights where SERN differs.
\subsection{Synchronizing physical and virtual domains with real time communication}

Prior work has moved toward tighter coupling between robots in the field and their virtual counterparts, but most systems still emphasize either simulation fidelity or operator interaction rather than continuous synchronization during execution. On the simulation side, RoboNetSim, ROS NetSim, and CORNET showed that multi robot experiments become far more realistic when robot autonomy and network behavior are modeled together \cite{kudelski2013robonetsim,calvo2021ros,acharyacornet2}. These frameworks are useful for studying communication effects and for testing coordination policies before deployment. However, they are primarily co simulation tools. They do not focus on maintaining a live shared state between physical robots and a virtual twin once the mission is underway. Another line of work has brought virtual and physical agents into the same operational loop through digital twins and mixed reality interfaces. Liang et al. \cite{liang2020bi} presented a bidirectional bridge between construction robots and their digital twin, showing the value of state synchronization across domains. Zheng et al. \cite{zheng2023integrated} combined virtual and physical swarm agents within a mixed reality interface, and Trout et al. \cite{trout2018collaborative} and Dennison et al. \cite{dennison2020creating} explored mixed reality environments for collaborative decision support and shared operational views. More recently, mixed reality digital twin work has shown how physical and simulated agents can be used together in hybrid sim to real experiments for multi agent learning \cite{samak2025mrdt}. These systems clearly show the promise of cross reality robotics, but they still leave open a practical systems question: how to keep real robots and the virtual environment synchronized through lightweight, selective message exchange that can support live multi robot execution rather than only training, visualization, or post hoc analysis. Unlike systems that use the virtual side mainly for simulation or visualization, our approach keeps the virtual environment actively coupled to the physical team during execution. It does this through a lightweight ROS based bridge that supports bidirectional updates, task level coordination, and efficient synchronization.

\subsection{Coordination, planning, and selective information sharing}

Multi robot coordination has long been shaped by communication limits as much as by planning itself. Early surveys and reviews made clear that reliable coordination depends on what information is shared, how often it is shared, and how robustly it travels across the network \cite{cao2013overview,yan2013survey,jawhar2018networking,chibani2021critical}. More recent work has improved coordination quality under uncertainty and scale. Stadler et al. \cite{stadler2023approximating} modeled the value of collaborative team actions in uncertain graphs, and SHARD improved persistent multi agent path finding with scalability and performance guarantees \cite{leet2022shard}. At the same time, recent exploration work under low bandwidth communication has shown that strong multi robot performance often depends on careful control of what is exchanged, because rich map sharing quickly becomes too expensive in constrained settings \cite{bayer2026lowbandwidth}. These studies are important because they move beyond ideal communication assumptions, but they still do not integrate a live virtual twin into the coordination loop. Although this line of work considers realistic communication constraints, it still does not integrate a live virtual twin into the coordination loop. SERN addresses this at the middleware level. Instead of forwarding all ROS traffic across domains, it prioritizes the state, control, and perception messages most relevant to coordinated execution. This keeps the virtual twin responsive and useful while limiting unnecessary communication overhead. In this way, SERN brings planning, and communication aware coordination together with an execution layer that keeps physical robots and the virtual environment connected in real time.
\section{Methodology}
This section describes the main components of SERN and how they work together under constrained communication.

\subsection{SERN System Overview}
\begin{figure*}[!htb]
    \centering
    \subfloat[]{
        \includegraphics[width=0.48\textwidth,height=4cm,keepaspectratio]{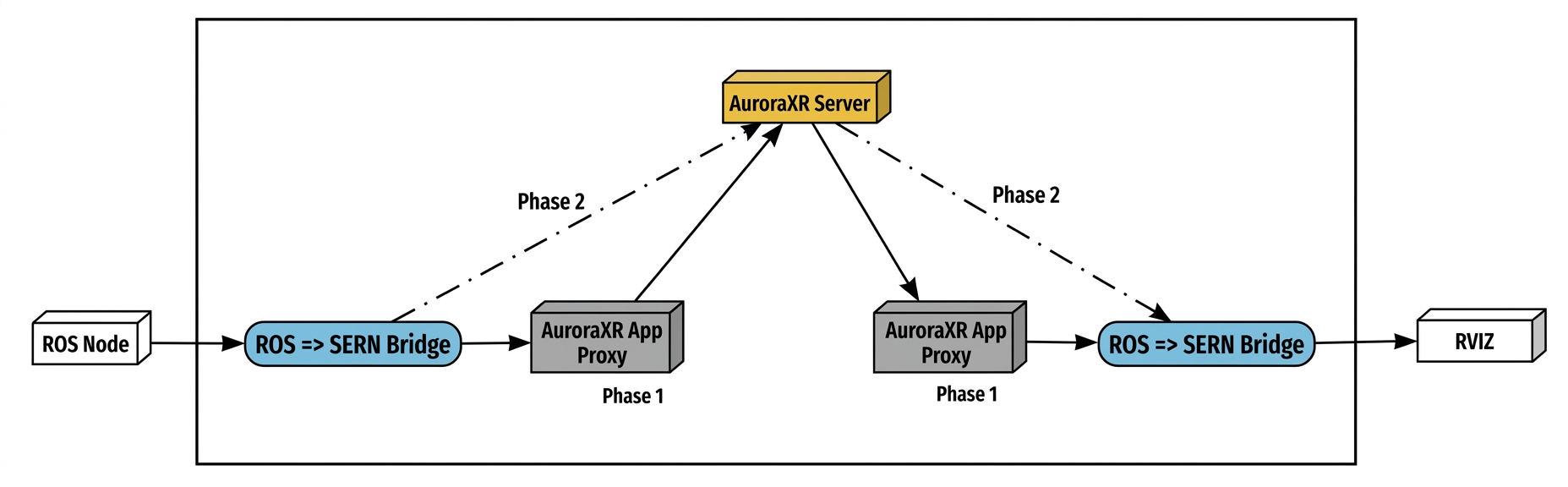}
        \label{fig:aurora_bridge}
    }
    \hfill
    \subfloat[]{
    \includegraphics[width=0.48\textwidth,height=4cm,keepaspectratio]{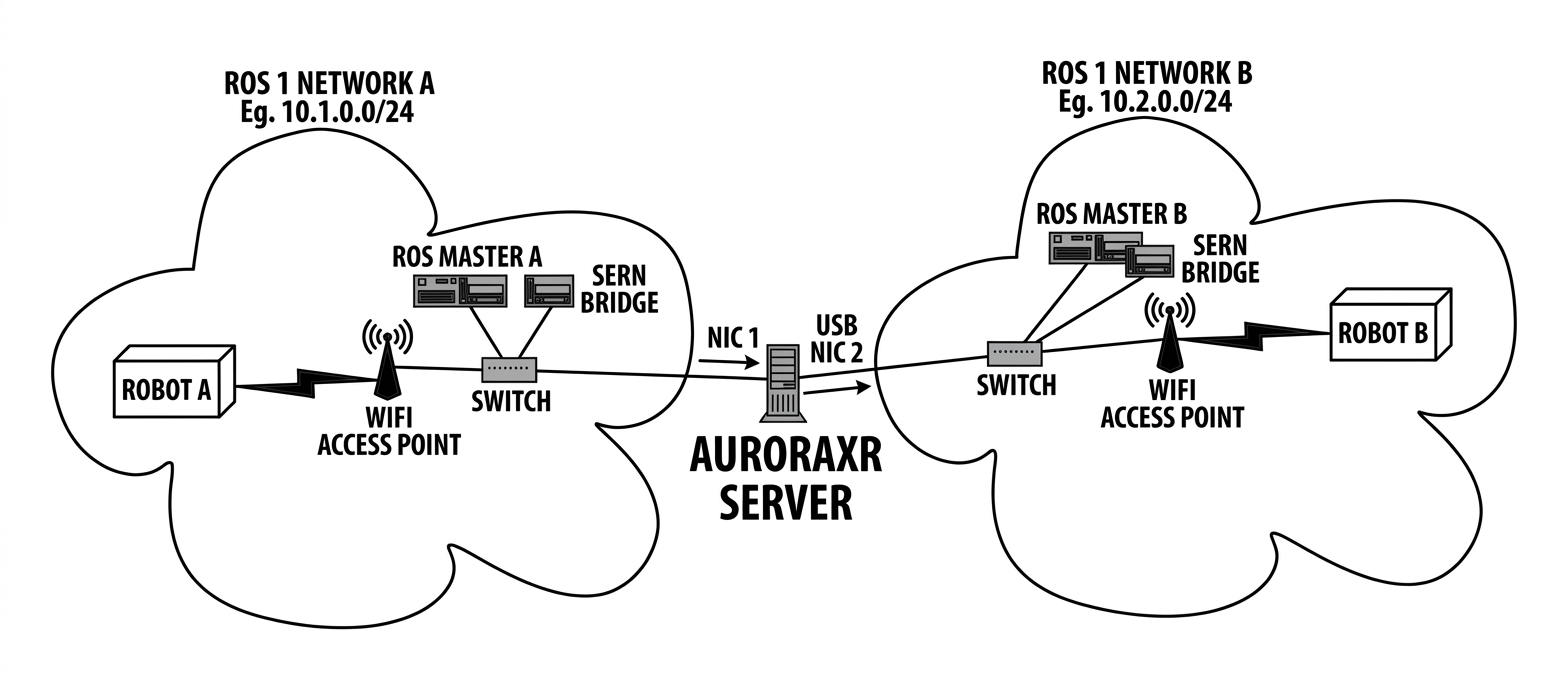}
        \label{fig:bridge_local_network}
    }
    \caption{SERN system overview. (a) Logical bridge architecture showing the cross reality communication and synchronization path. (b) Physical deployment and network topology used to connect distributed ROS systems through the central server.}
    \label{fig:sern_overview}
\end{figure*}

SERN is organized around one practical goal: keep a simulation side twin useful during live robot operation when communication is delayed, lossy, or bandwidth limited. Figure~\ref{fig:sern_overview} shows the two parts of the system. Figure~\ref{fig:aurora_bridge} gives the logical data path from ROS topics to the cross reality server and back to a remote ROS endpoint. Figure~\ref{fig:bridge_local_network} shows the deployed network layout used in our experiments. At runtime, the physical robot publishes telemetry, control state, and selected perception outputs through the \textbf{SERN ROS Bridge}. The bridge forwards those streams to the server, which redistributes them to the virtual twin and to remote ROS endpoints. In the other direction, the same path carries configuration changes, operator inputs, and visualization updates. SERN combines this communication path with a synchronization loop that predicts the twin state during communication gaps and corrects it when fresh telemetry arrives.
The architecture can relay multiple ROS namespaces, but the field validation uses \emph{one physical robot}. Additional namespaces in the scaling tests are replayed ROS streams used to increase communication load and evaluate bridge behavior.

\subsection{Representing the Real Environment in the Virtual Twin}
\label{sec:virtual_twin}
The virtual twin must satisfy three requirements. First, it must stay geometrically consistent with the physical site. Second, it must accept live updates from robot telemetry with low delay. Third, it must remain light enough to update in real time when compute and bandwidth are limited. We build the twin from static map layers and live robot data. Static sources, such as digital elevation maps and vector layers, provide the base terrain and scene layout. Live sources, such as odometry, IMU, and LiDAR, provide local updates that reflect changes seen by the robot during operation.

We represent these live updates as \emph{scene patches}. Each patch describes only the local part of the scene that has changed since the last synchronized state, such as a new obstacle region, a corrected pose estimate, or a local occupancy change. This keeps the update traffic smaller than sending a full scene refresh every cycle. In our implementation, the robot state is localized in a local tangent frame, the incoming LiDAR and odometry are fused into an occupancy representation, and only the difference between the current twin state and the new estimate is sent to Unity. The resulting patch stream is applied at 10\,Hz during normal operation.

\noindent \textbf{Geodesic to local to Unity transform.}
We use a geodetic reference point $(\phi_0,\lambda_0,h_0)$ to map GPS measurements into a local metric frame before sending them to Unity. For larger sites we use the Haversine distance
\begin{equation}
 d \,=\, 2r\arcsin\!\left(\sqrt{\sin^2\!\frac{\Delta\phi}{2} + \cos\phi_1\cos\phi_2\sin^2\!\frac{\Delta\lambda}{2}}\right),
\label{eq:haversine}
\end{equation}
where $r$ is Earth's radius, latitude and longitude are in radians, and $d$ is in meters. For the smaller field sites used in our experiments, a local tangent approximation is sufficient:
\begin{equation}
\Delta x = r\cos\phi_0\,\Delta\lambda,\qquad
\Delta y = r\,\Delta\phi,\qquad
\Delta z = h-h_0.
\label{eq:enu}
\end{equation}
The Unity coordinates are then obtained from
\begin{equation}
\mathbf{u} = s\,\mathbf{R}\,[\Delta x,\Delta y,\Delta z]^\top,
\label{eq:unity}
\end{equation}
where $\mathbf{R}\in SO(3)$ encodes the axis convention used by the Unity scene and $s>0$ maps meters into engine units (Alg.~\ref{alg:convert_gps_to_unity}). 

\noindent \textbf{Patch based update policy.}
Not every part of the world needs the same update rate. Areas near the robot, the current route, or task relevant landmarks are updated at higher fidelity than distant static regions. Let $\mathcal{C}$ denote the set of currently important regions. For each cell or mesh block $v$, we assign one of three update levels (Alg.~\ref{alg:lod_hyst}),
\begin{equation}
\mathrm{LoD}(v)=
\begin{cases}
L_{\text{high}}, & v\in\mathcal{C},\\
L_{\text{med}}, & \mathrm{prox}(v,\mathcal{C})<\rho_{\downarrow},\\
L_{\text{low}}, & \text{otherwise,}
\end{cases}
\label{eq:lod}
\end{equation}
with hysteresis thresholds $\rho_{\downarrow}<\rho_{\uparrow}$ to avoid rapid switching near region boundaries. This policy reduces unnecessary updates outside the robot's active area while keeping the local scene responsive.

\begin{algorithm}
\caption{Geo to Local to Unity Transform (batched)}
\label{alg:convert_gps_to_unity}
\begin{algorithmic}[1]
\Require Reference $(\phi_0,\lambda_0,h_0)$; batch $\{(\phi_k,\lambda_k,h_k)\}$ in radians and meters; scale $s$; axis map $\mathbf{R}$
\ForAll{$k$ \textbf{in parallel}}
  \State $\Delta\phi \gets \phi_k-\phi_0$, $\Delta\lambda \gets \lambda_k-\lambda_0$
  \State $\Delta x \gets r\cos\phi_0\,\Delta\lambda$, $\Delta y \gets r\,\Delta\phi$, $\Delta z \gets h_k-h_0$
  \State $\mathbf{u}_k \gets s\,\mathbf{R}[\Delta x,\Delta y,\Delta z]^\top$
\EndFor
\State \Return $\{\mathbf{u}_k\}$
\end{algorithmic}
\end{algorithm}

\begin{algorithm}
\caption{Patch Update Level Selection with Hysteresis}
\label{alg:lod_hyst}
\begin{algorithmic}[1]
\Require Grid $G$, important region set $\mathcal{C}$, thresholds $(\rho_\downarrow,\rho_\uparrow)$
\ForAll{$v\in G$ \textbf{in parallel}}
  \State $d \gets \mathrm{prox}(v,\mathcal{C})$
  \If{$v\in\mathcal{C}$}
      \State $\mathrm{LoD}(v) \gets L_{\text{high}}$
  \ElsIf{$d<\rho_\downarrow$ \textbf{or} $(d<\rho_\uparrow$ \textbf{and} $\mathrm{LoD}(v)\neq L_{\text{low}})$}
      \State $\mathrm{LoD}(v) \gets L_{\text{med}}$
  \Else
      \State $\mathrm{LoD}(v) \gets L_{\text{low}}$
  \EndIf
\EndFor
\end{algorithmic}
\end{algorithm}

\subsection{Distribution of ROS Data Under Constrained Communication}
\label{sec:virtual_twin_distribution}
Standard ROS communication works well inside a stable local network, but it becomes fragile when the link between the robot and the remote twin is delayed, lossy, or bandwidth limited. SERN addresses this by extending the AuroraXR cross reality framework~\cite{dennison2022auroraxr} with a custom \textbf{SERN ROS Bridge}. The bridge is designed around three simple ideas.

\textit{(1) Priority first transmission.} Topics that are directly tied to safe control and state synchronization are sent first. In our implementation, command messages, robot pose, velocity, and timing information form the highest priority class. Map corrections, obstacle summaries, and selected perception outputs form a second class. Heavy or non essential streams, such as dense debug outputs, remain lowest priority. This ordering keeps the bridge focused on the information that matters most when the link is under stress.

\textit{(2) A common cross reality representation.} ROS topics, transforms, and selected node metadata are wrapped as AuroraXR objects so that they can move through the same server side path and be consumed by the twin, by remote ROS nodes, or by operator interfaces. This keeps the bridge logic separate from any one user interface and allows the same transport layer to support visualization, telemetry, and control~\cite{quigley2009ros,dennison2022auroraxr}.

\textit{(3) Compact state exchange.} When bandwidth is limited, the bridge favors compact, task relevant summaries over raw high bandwidth streams whenever an equivalent summary is available. For example, the navigation stack can send obstacle masks, occupancy updates, and robot pose estimates rather than forwarding every raw point cloud frame. The goal is not to remove sensing detail everywhere, but to preserve the information needed for synchronization and navigation with a smaller communication footprint.

To implement the first item above, the bridge uses a bandwidth-aware scheduler (\textit{Alg.~\ref{alg:prioritize}} details this process). Each message inherits a topic criticality score $\kappa(t)$, and the scheduler combines that score with message size and waiting time. Let $m$ denote a pending message from topic $t$, let $s(m)$ be its estimated serialized size, and let $a(m)$ be its age in the queue. We assign the score
\begin{equation}
P(m)=\frac{\kappa(t)\,[1+\eta a(m)]}{s(m)+\varepsilon},
\label{eq:priority_score}
\end{equation}
where $\eta\ge 0$ controls how quickly waiting messages rise in priority and $\varepsilon>0$ avoids division by zero. This form favors compact, high-value messages, while the waiting time term helps prevent messages from being deferred for too many scheduling cycles. Given an available bandwidth budget $B_{\text{avail}}$, the scheduler transmits messages in decreasing order of $P(m)$ until the current interval budget is exhausted. Messages from the lowest priority class may be deferred or dropped if they become stale.

\begin{algorithm}
\caption{Bandwidth Aware ROS Message Prioritization}
\label{alg:prioritize}
\begin{algorithmic}[1]
\Require Pending message list $M$; topic criticality scores $\kappa(t)$; size function $s(m)$; age function $a(m)$; available bandwidth $B_{\text{avail}}$
\For{each message $m \in M$ from topic $t$}
    \State Compute $P(m) \gets \dfrac{\kappa(t)[1+\eta a(m)]}{s(m)+\varepsilon}$
\EndFor
\State Sort $M$ by $P(m)$ in descending order
\State $B_{\text{used}} \leftarrow 0$
\For{each message $m$ in sorted order}
    \If{$B_{\text{used}} + s(m) \le B_{\text{avail}}$}
        \State transmit $m$
        \State $B_{\text{used}} \leftarrow B_{\text{used}} + s(m)$
    \ElsIf{$m$ is stale and low priority}
        \State drop $m$
    \Else
        \State keep $m$ in queue for the next interval
    \EndIf
\EndFor
\end{algorithmic}
\end{algorithm}

\begin{figure}[!htb]
\centering
\includegraphics[width=\linewidth]{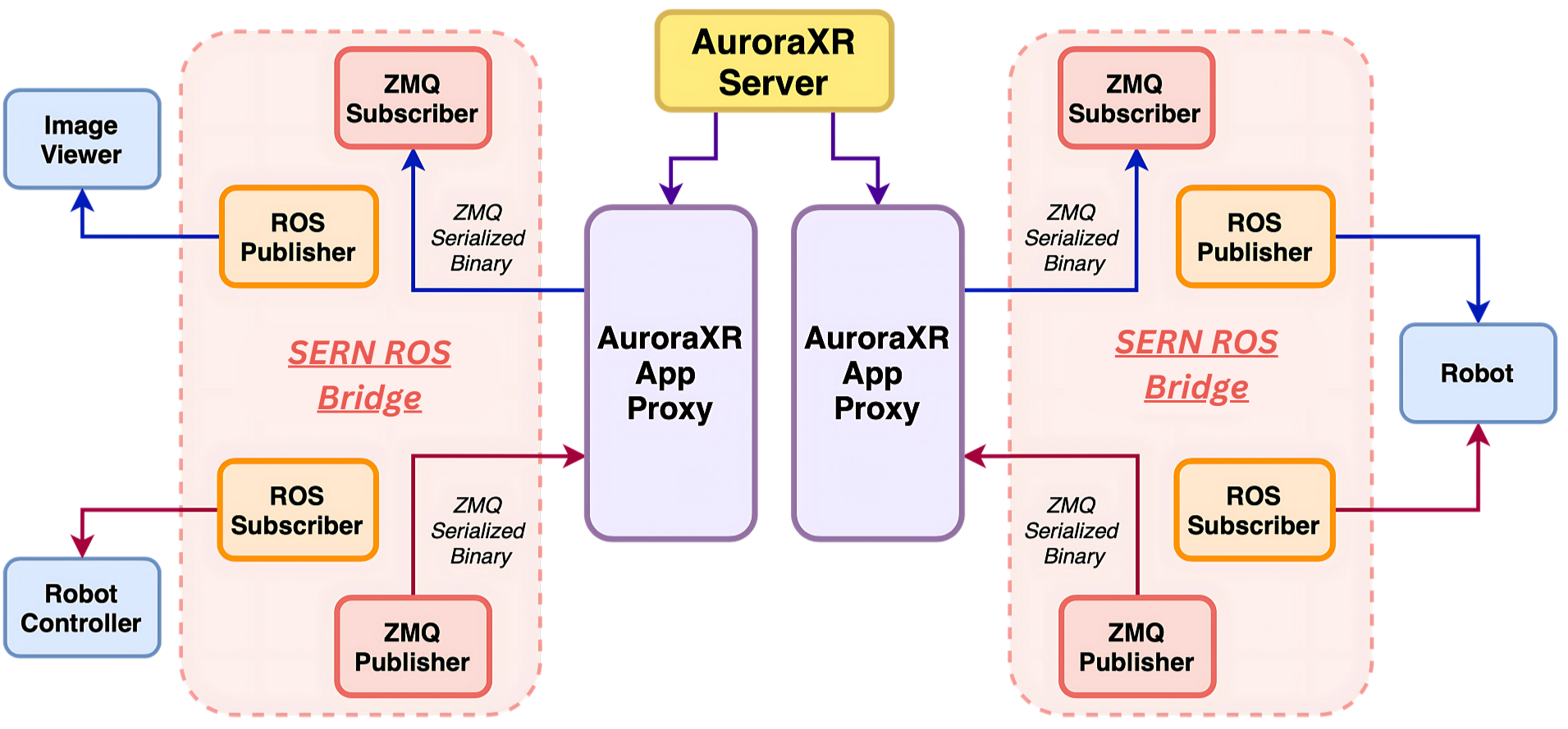}
\caption{Bi directional communication path used by the SERN ROS Bridge. Robot side ROS topics are serialized, forwarded through the AuroraXR server, and republished on the receiving side.}
\label{fig:communication_framework}
\end{figure}

\subsection{SERN ROS Bridge Implementation Details}

The SERN ROS Bridge is the bi-directional communication backbone connecting distributed ROS nodes with the virtual environment. It leverages an AuroraXR server as a secure relay for data transfer between networks. Fig.~\ref{fig:communication_framework} illustrates the architecture, which comprises several concurrent components to handle real-time data flow. At each robot site, a \emph{Bridge Client} runs with ROS publishers/subscribers and ZeroMQ sockets. Outgoing ROS messages are picked up by a ZMQ \textit{Publisher thread}, serialized into a binary format, and streamed to the AuroraXR server. On the receiving side (e.g., a remote operations center or another robot network), a ZMQ \textit{Subscriber thread} listens to the AuroraXR feed, deserializes incoming messages, and republishes them on the local ROS topic space. This effectively mirrors ROS topics across network boundaries. From the point of view of the application, this makes selected topics appear on both sides of the connection even though they reside on different networks.

The bridge supports two configuration modes. In \textbf{manual mode}, the operator specifies which topics should be relayed and assigns each topic to a priority class. This is the mode used for the main field trials because it gives full control over what crosses the link. In \textbf{dynamic discovery mode}, the bridge monitors the ROS master and adds newly observed topics at runtime. Dynamic discovery is an implementation convenience that allows the system to incorporate new namespaces without restarting the bridge. On top of the basic relay path, the bridge maintains per topic queues, timestamps transmitted messages, and keeps a small replay buffer for selected state streams. The replay buffer is useful when a short dropout ends and the receiving side needs the most recent state without waiting for the next full update cycle. The bridge also records message timing statistics that are later used by the runtime configuration logic described in Section~\ref{sec:multi-metric-cost}.

\subsection{Physics Aware Synchronization between the Physical Robot and the Twin}
\label{sec:physics-aware-sync}
Reliable navigation requires the twin to remain close enough to the physical robot to support planning and monitoring, even when telemetry is delayed or temporarily missing. A simple copy whenever data arrive is not enough because the virtual state can become stale during communication gaps. SERN addresses this with a predictor correction loop.

Let the synchronized state at time step $k$ be $x_k=[p_k,v_k]^\top$, where $p_k\in\mathbb{R}^n$ is position and $v_k\in\mathbb{R}^n$ is velocity. During normal operation, the twin is updated directly from incoming telemetry. During a communication gap, the twin is propagated using the most recent motion estimate. We write this prediction as
\begin{align}
\hat{v}_{k+1} &= \hat{v}_k + a_k\,\Delta t, \\
\hat{p}_{k+1} &= \hat{p}_k + \hat{v}_k\,\Delta t + \tfrac{1}{2}a_k\,\Delta t^2,
\label{eq:prediction_step}
\end{align}
where $a_k$ is an acceleration estimate obtained from the recent IMU and odometry history together with the last issued motion command. In practice, this prediction is executed by the Unity side rigid body model, while the equations above describe the state update used by the bridge side synchronizer.

When fresh telemetry returns, we compare the measured physical state $x_k^{\text{phys}}$ with the current virtual state $x_k^{\text{virt}}$ and compute the synchronization error
\begin{equation}
e_k = x_k^{\text{phys}} - x_k^{\text{virt}} = \begin{bmatrix} e_{p,k} \\ e_{v,k} \end{bmatrix}.
\label{eq:error_state}
\end{equation}
The bridge then applies a smooth correction to the virtual side rather than snapping the state immediately. We use a PD style correction term
\begin{equation}
u_{\text{corr},k} = K_p(\tau_k)\,e_{p,k} + K_d(\tau_k)\,e_{v,k},
\label{eq:pd_correction}
\end{equation}
where $\tau_k$ is the time since the last reliable telemetry update. To avoid aggressive corrections after a long dropout, the gains are reduced as the outage duration grows,
\begin{equation}
\begin{aligned}
K_p(\tau) &= \mathrm{clip}\!\left(K_p^0 e^{-\lambda_p \tau},\,K_p^{\min},\,K_p^{\max}\right), \\
K_d(\tau) &= \mathrm{clip}\!\left(K_d^0 e^{-\lambda_d \tau},\,K_d^{\min},\,K_d^{\max}\right).
\end{aligned}
\label{eq:adaptive_gains}
\end{equation}
so that short gaps are corrected quickly while longer gaps are corrected more smoothly. The same outage duration also controls a tolerance band $\epsilon(\tau)=\epsilon_0+\alpha_\epsilon\tau$ used to decide when a corrective update is necessary.

\noindent \textbf{Bounded growth during a communication gap.}
The prediction model above gives a simple bound on how far the twin can diverge during an outage. Suppose the acceleration mismatch between the physical and predicted motion is bounded by $\|a^{\text{phys}}(t)-a^{\text{pred}}(t)\|\le \bar{a}$ for $t\in[t_0,t_0+\tau]$. Then the velocity and position errors satisfy
\begin{align}
\|e_v(t_0+\tau)\| &\le \|e_v(t_0)\| + \bar{a}\,\tau, \\
\|e_p(t_0+\tau)\| &\le \|e_p(t_0)\| + \|e_v(t_0)\|\tau + \tfrac{1}{2}\bar{a}\,\tau^2.
\label{eq:gap_bound}
\end{align}
This bound does not claim perfect tracking under arbitrarily long outages. It simply shows that, over a finite gap with bounded motion mismatch, the prediction error grows in a controlled way rather than instantaneously diverging. Once telemetry resumes, the correction term in~\eqref{eq:pd_correction} pulls the virtual state back toward the physical state.

To make the synchronization behavior more explicit during temporary communication loss, we use a finite interval error bound over the gap duration.

\begin{proposition}
\label{prop:finite_horizon_gap}
Let $x_p(t)$ and $x_v(t)$ denote the physical and virtual robot states, and suppose the dynamics satisfy
\[
\|f(x_1,u_1)-f(x_2,u_2)\|
\le L_x \|x_1-x_2\| + L_u \|u_1-u_2\|
\]
for some constants $L_x,L_u>0$. Define the synchronization error as
\[
e(t)=\|x_p(t)-x_v(t)\|.
\]
Assume that over a communication gap interval $t\in[t_0,t_0+T]$, the control mismatch is bounded as
\[
\|u_p(t)-u_v(t)\|\le \delta_{\max}.
\]
Then for all $t\in[t_0,t_0+T]$,
\[
e(t)\le e(t_0)e^{L_x(t-t_0)}
      + \frac{L_u}{L_x}\bigl(e^{L_x(t-t_0)}-1\bigr)\delta_{\max}.
\]
In particular, if the systems are synchronized at the start of the gap so that $e(t_0)=0$, then
\[
e(t)\le \frac{L_u}{L_x}\bigl(e^{L_x(t-t_0)}-1\bigr)\delta_{\max}.
\]
\end{proposition}

\noindent\textit{Proof.}
From
\[
\dot x_p = f(x_p,u_p), \qquad \dot x_v = f(x_v,u_v),
\]
we have
\[
\frac{d}{dt}(x_p-x_v)=f(x_p,u_p)-f(x_v,u_v).
\]
Using the Lipschitz bound and the definition of $e(t)$,
\[
\dot e(t)\le L_x e(t)+L_u\|u_p(t)-u_v(t)\|
          \le L_x e(t)+L_u\delta_{\max}.
\]
Applying Gr\"onwall's inequality over $[t_0,t]$ gives
\[
e(t)\le e(t_0)e^{L_x(t-t_0)}
      + \int_{t_0}^t L_u\delta_{\max} e^{L_x(t-\tau)}\,d\tau,
\]
which evaluates to
\[
e(t)\le e(t_0)e^{L_x(t-t_0)}
      + \frac{L_u}{L_x}\bigl(e^{L_x(t-t_0)}-1\bigr)\delta_{\max}.
\]
\hfill$\square$

This finite interval bound is consistent with the behavior observed in the 30\,s communication gap experiment ( Table~\ref{tab:sync_summary}), where predictive propagation kept the virtual state close enough to the physical robot for smooth correction once updates resumed.

\subsection{Multi-Metric Cost Function for Bridge Configuration}
\label{sec:multi-metric-cost}
The bridge exposes several runtime choices that trade communication quality against resource use. We collect these choices into a configuration vector
\begin{equation}
\mathbf{c} = [r_1,\ldots,r_q,\, q_1,\ldots,q_q,\, z,\, h]^\top,
\label{eq:config_vector}
\end{equation}
where $r_i$ is the rate cap or downsampling factor for topic class $i$, $q_i$ is the queue limit for that class, $z$ is the compression mode, and $h$ denotes whether extra redundancy is enabled for selected high priority topics. The bridge does \emph{not} use MMCF to tune the synchronization gains in Section~\ref{sec:physics-aware-sync}. MMCF is only used to choose communication behavior.

For each candidate configuration $\mathbf{c}$, we measure four normalized quantities over a recent time window: latency $L(\mathbf{c})$, loss $P(\mathbf{c})$, compute cost $C(\mathbf{c})$, and bandwidth use $B(\mathbf{c})$. We define
\begin{equation}
L(\mathbf{c}) = \frac{\ell(\mathbf{c})-\ell_{\min}}{\ell_{\max}-\ell_{\min}},
\qquad
P(\mathbf{c}) = \frac{p(\mathbf{c})-p_{\min}}{p_{\max}-p_{\min}},
\label{eq:lat_loss_norm}
\end{equation}
and
\begin{equation}
C(\mathbf{c}) = \frac{\tau(\mathbf{c})}{\tau_{\max}},
\qquad
B(\mathbf{c}) = \frac{b(\mathbf{c})}{b_{\max}},
\label{eq:comp_band_norm}
\end{equation}
where $\ell(\mathbf{c})$ is the measured end to end latency, $p(\mathbf{c})$ is the packet loss rate, $\tau(\mathbf{c})$ is the bridge side processing time, and $b(\mathbf{c})$ is the observed bandwidth use. The corresponding cost is
\begin{equation}
\mathrm{MMCF}(\mathbf{c}) = \alpha L(\mathbf{c}) + \beta P(\mathbf{c}) + \gamma C(\mathbf{c}) + \delta B(\mathbf{c}),
\label{eq:mmcf}
\end{equation}
with nonnegative weights satisfying $\alpha+\beta+\gamma+\delta=1$.

Every configuration interval $T_{\text{cfg}}$, the bridge evaluates a finite set of feasible configurations and selects
\begin{equation}
\mathbf{c}^* = \arg\min_{\mathbf{c}\in\mathcal{C}_{\text{feas}}} \mathrm{MMCF}(\mathbf{c}).
\label{eq:mmcf_argmin}
\end{equation}
The selected configuration is then applied for the next interval. In a latency sensitive profile, the weights place more emphasis on $L$ and $P$. In a bandwidth limited profile, they place more emphasis on $B$ while keeping enough weight on $P$ to avoid unreliable operation. This makes the bridge behavior explicit: MMCF chooses how aggressively to downsample, queue, compress, and protect traffic under the current network condition.

\subsection{Environmental Perception and Navigation Planning}
Beyond state synchronization, our framework integrates perception and planning capabilities to validate that the cross-reality system can help real-world decision-making. We deploy a \textbf{semantic perception module} in both the physical and virtual domains to classify and label environmental features (ground, obstacles, etc.). In the virtual world, we leverage labeled 3D models and simulated LiDAR scans to perform semantic segmentation, which helps the virtual agent identify critical objects. In the physical domain, the robot uses 3D LiDAR and on-board vision to perform semantic segmentation of its surroundings. To bridge the gap between the different domains and sensors, we apply a domain adaptation technique ~\cite{wu2019squeezesegv2} that aligns the feature distributions of virtual and real sensor data, improving the model’s generalization without requiring extensive re-training on the real data. For example, a tree log lying across a virtual road is recognized as an obstacle in simulation, and the system can similarly recognize real debris or objects in the physical environment, even if they were not in the training set.

For navigation, the robot uses a hybrid of virtual and physical information. We convert 3D point clouds to 2D occupancy grid slices for efficient path planning: given the LiDAR point cloud $L_{3D}(r,\theta,z)$ (range, angle, height), we compute a 2D obstacle map 
$S_{2D}(r,\theta) = \min_{z} L_{3D}(r,\theta,z)$, 
which essentially collapses the vertical dimension by taking the lowest point (highest obstruction) at each bearing $(r,\theta)$.  A SLAM module then uses this 2D scan to update a map and plan a path to the goal while avoiding obstacles. Importantly, the robot cross-checks any virtual obstacles with its real sensor data: if the virtual twin observes an obstacle that the physical sensors do not detect (or vice versa), the robot flags the discrepancy and adjusts its plan accordingly (favoring its own sensor data for safety). 

\begin{figure*}[!htb]
    \centering
    \subfloat[]{%
        \includegraphics[width=0.32\textwidth,height=3cm]{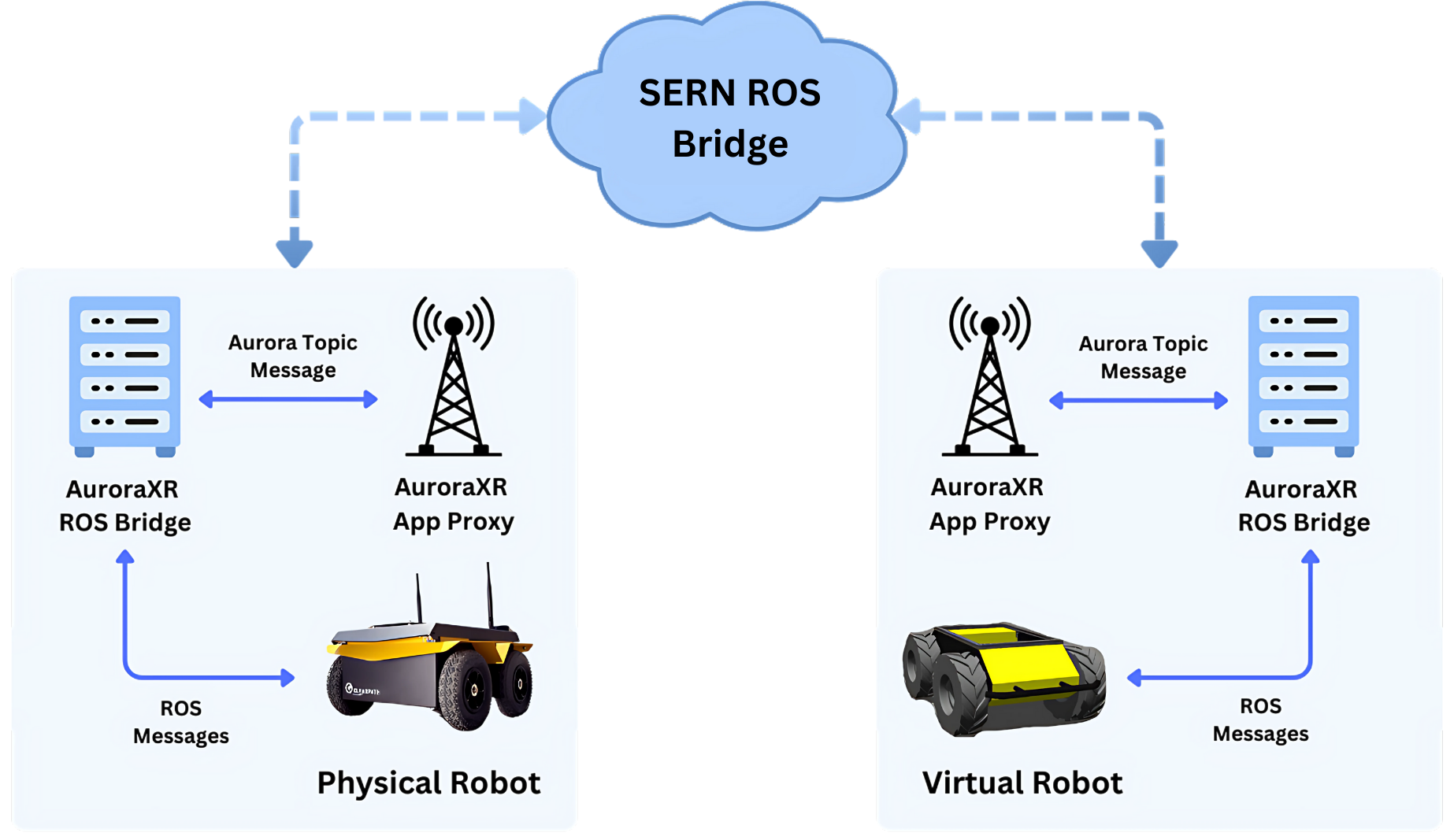}
        \label{fig:aurora_bridge_topology}
    }
    \hfill
    \subfloat[]{%
        \includegraphics[width=0.32\textwidth,height=3cm]{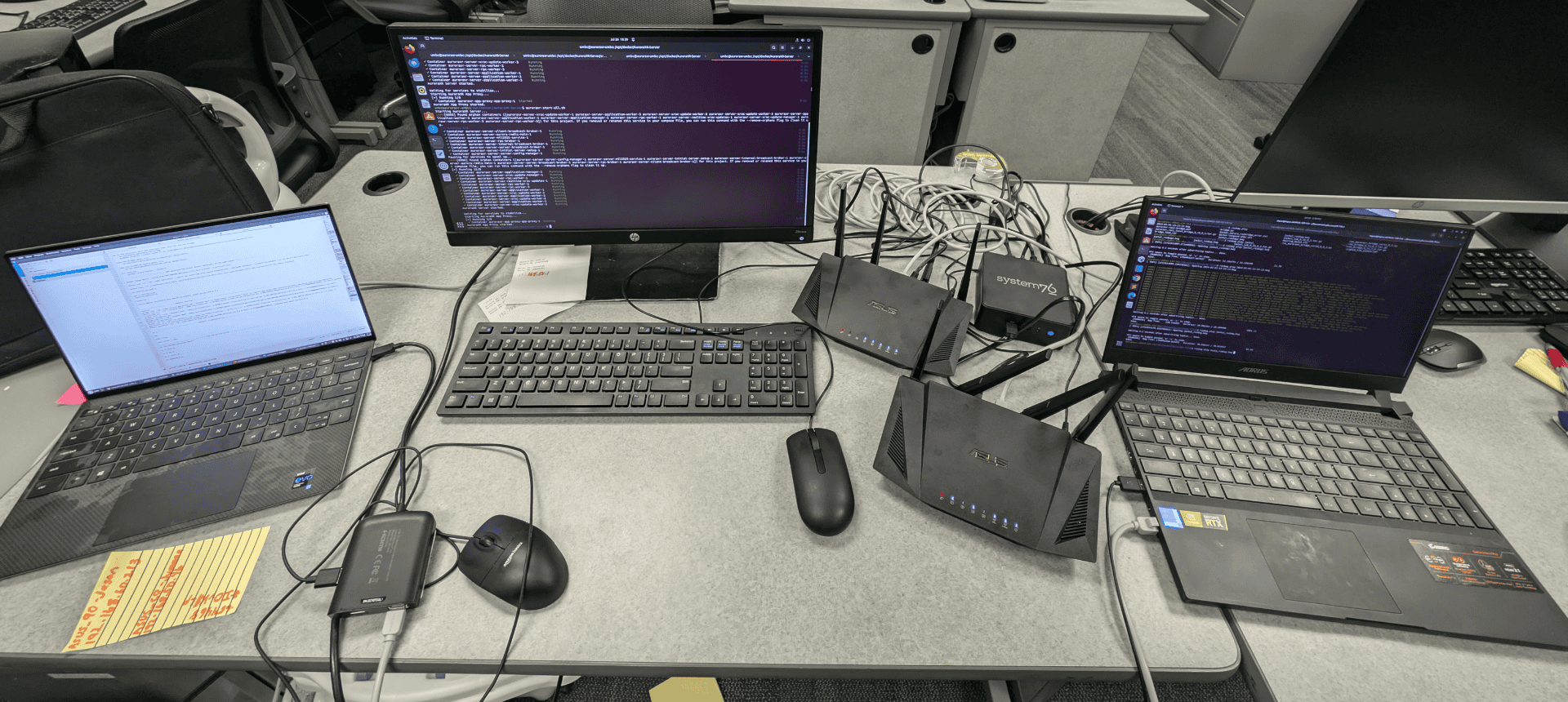}
        \label{fig:local_network_setup}
    }
    \hfill
    \subfloat[]{%
        \includegraphics[width=0.32\textwidth,height=3cm]{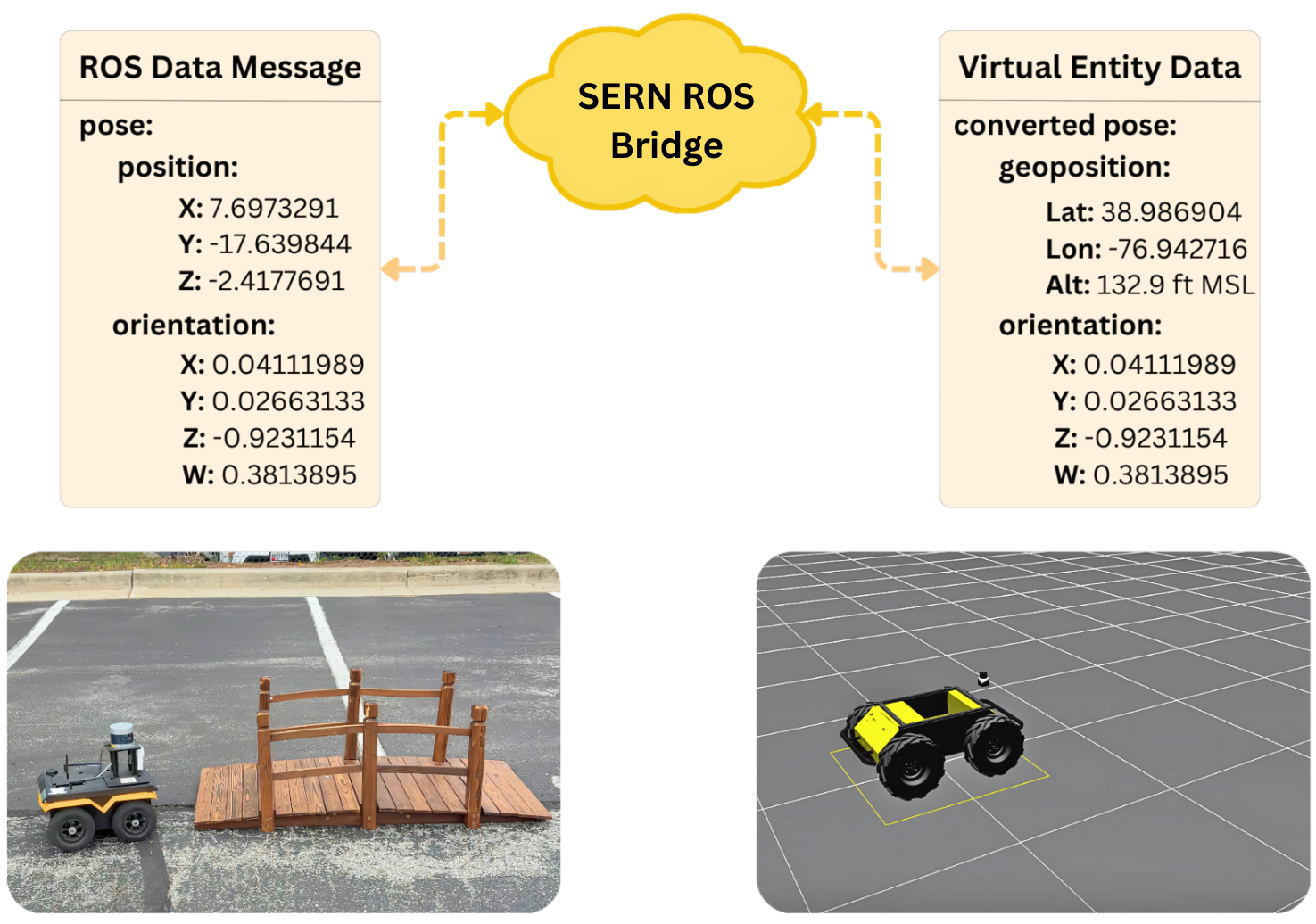}
        \label{fig:receiver_visualization}
    }
   \caption{SERN ROS Bridge deployment and testbed. (a) Logical relay path used to move ROS data across separate subnets through the SERN server, showing how robot side topics are discovered, forwarded, and made available to remote clients. (b) Hardware and network testbed used in the bridge evaluation, including robot side ROS hosts, local switches, WiFi access points, and the central SERN server with multiple network interfaces connecting the two ROS subnetworks. (c) Receiver side visualization of bridged data in RViz, showing robot state, laser scans, and map updates after transmission through the bridge.}
    \label{fig:auroraxr-setup}
\end{figure*}

\begin{figure*}[!htb]
    \centering
    \subfloat[]{%
        \includegraphics[width=0.24\textwidth,height=3cm]{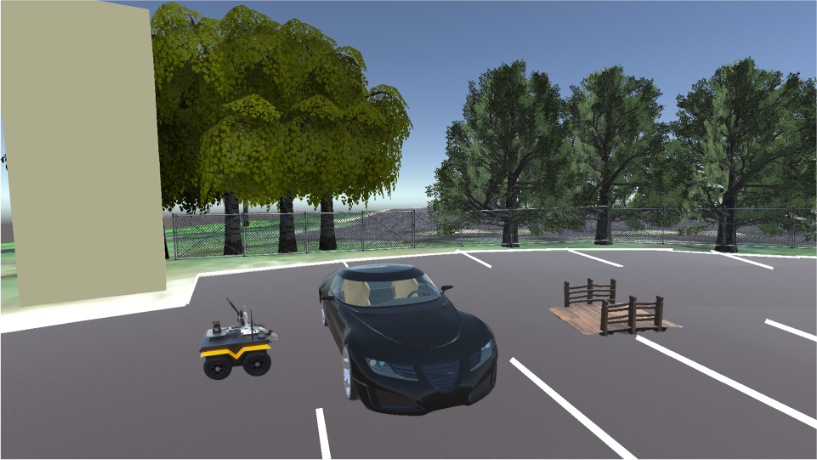}
   }
   \hfill
    \subfloat[]{%
        \includegraphics[width=0.24\textwidth,height=3cm]{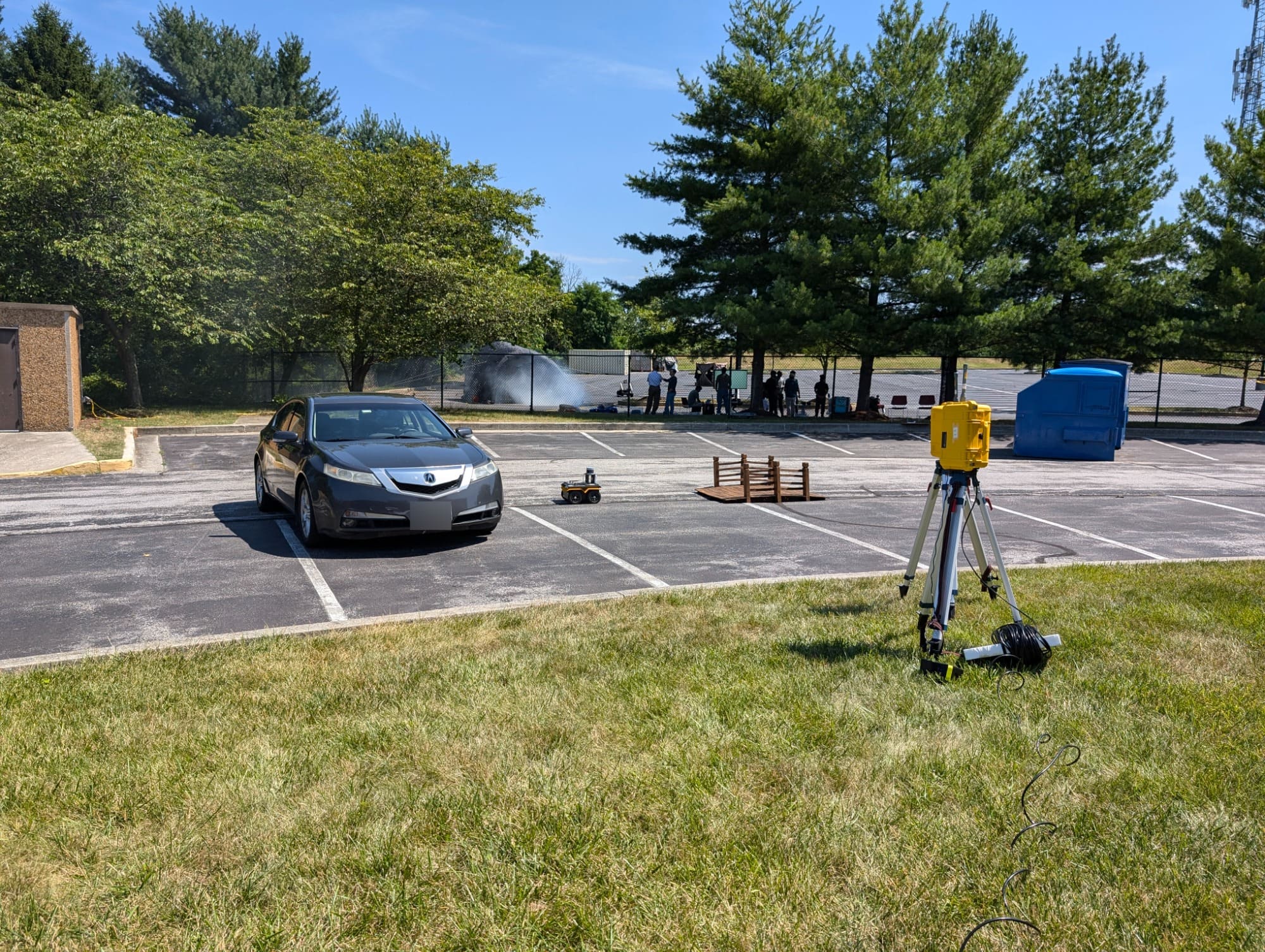}
    }
    \hfill
    \subfloat[]{%
        \includegraphics[width=0.24\textwidth,height=3cm]{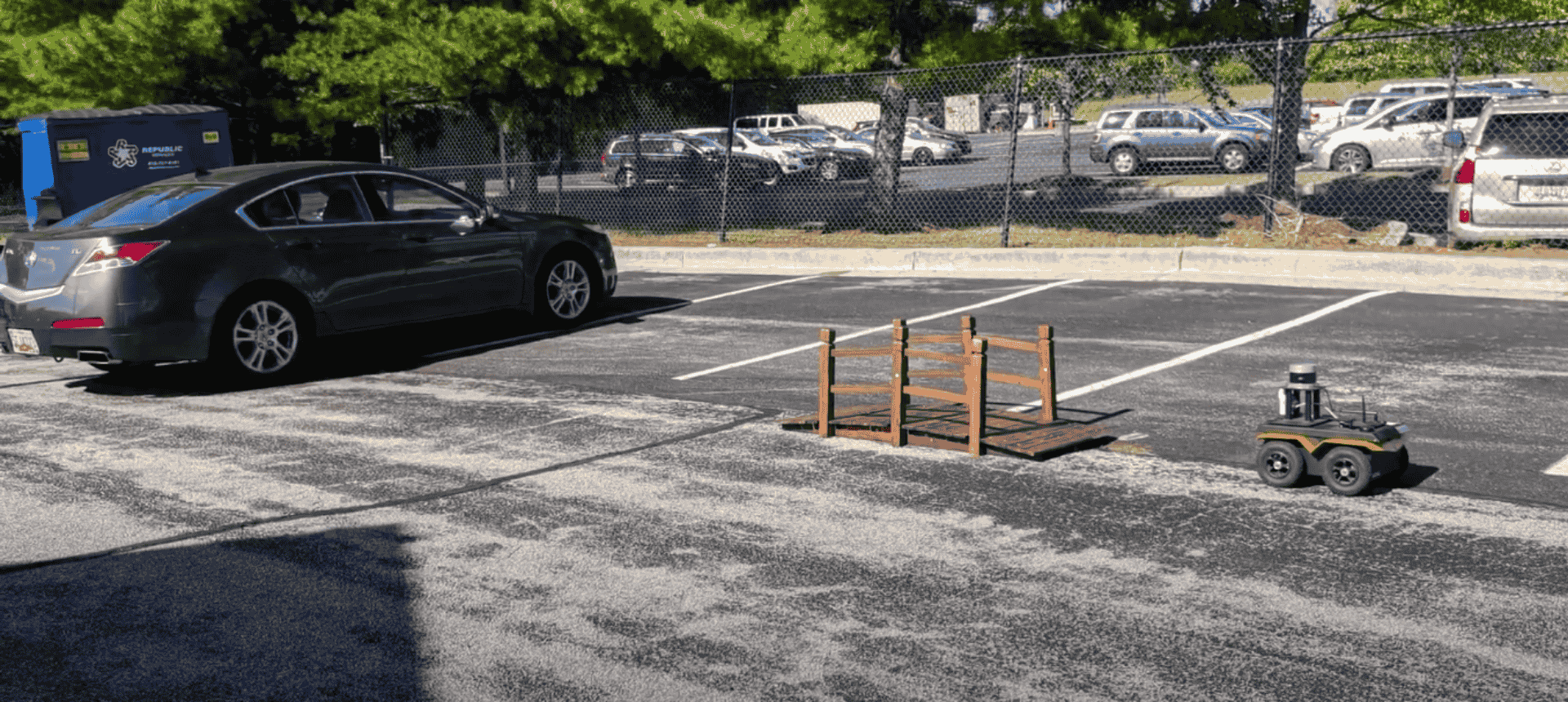}
    }
    \hfill
    \subfloat[]{%
        \includegraphics[width=0.24\textwidth,height=3cm]{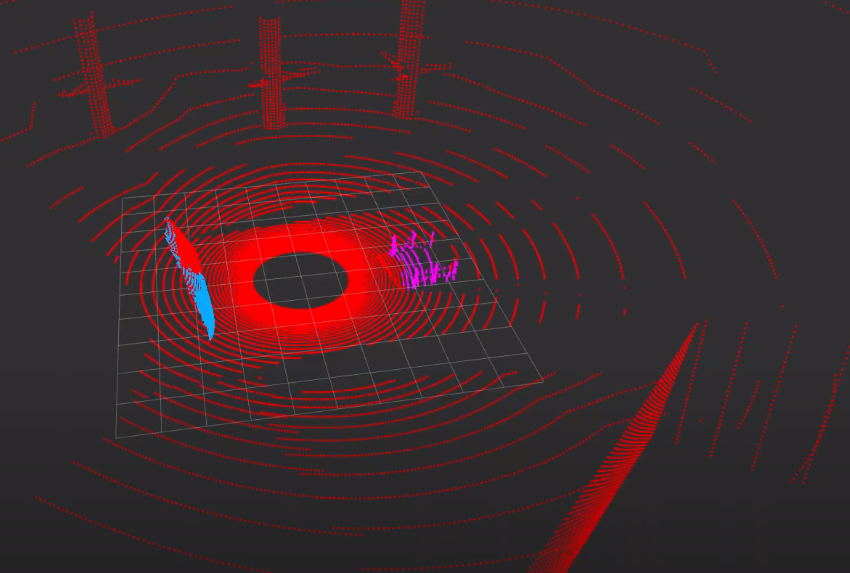}
        \label{fig:semantic_segmentation}
    }
    \caption{Virtual and physical settings used in the navigation study. (a) Simulation side environment. (b) Physical site under degraded communication. (c) Field navigation trial. (d) Example low bandwidth obstacle representation used by the twin and the robot.}
    \label{fig:virtual_physical_navigation}
\end{figure*}

\section{Experiments}
\label{experiment_result}

This section evaluates SERN as a cross reality synchronization and communication framework for simulation enhanced robot navigation. We focus on four questions: (i) whether SERN keeps the virtual twin aligned with the physical robot under constrained communication, (ii) whether the bridge reduces communication overhead relative to a standard ROS setup, (iii) whether the combined real and virtual pipeline improves navigation performance, and (iv) whether the multi metric cost function selects lower cost bridge settings than fixed policies.

\subsection{Experimental Setup}

We evaluate SERN in a cross reality navigation setting built around one physical robot and a simulation side twin. The field experiments use a single physical robot for synchronization and navigation validation. To study bridge behavior under higher communication load, we replay recorded ROS bag streams under separate namespaces on a laptop. These replayed streams are used to stress the communication layer and evaluate scaling behavior of the bridge. 

The testbed consists of two ROS networks connected through an AuroraXR relay server and the SERN ROS Bridge. One network hosts the physical robot and the second network hosts the virtual side twin and monitoring tools. Figure~\ref{fig:auroraxr-setup} shows the deployment. The central AuroraXR server routes selected ROS topics between the two networks while keeping the networks separate. The physical platform is a Clearpath Jackal or Husky class ground robot equipped with onboard sensing, and the virtual side runs in Unity with a matching robot instance. The server runs on a standard desktop with an Intel i7 CPU at 2.8~GHz and 16~GB RAM. The Unity side runs on a laptop with an NVIDIA RTX~3090 GPU. Unless noted otherwise, the two ROS networks communicate over local WiFi with a typical round trip latency of 5--10~ms.


We designed a representative cross-reality navigation scenario to demonstrate SERN in action. We built a high-fidelity virtual twin of a campus environment using Unity, including terrain and structural features identical to the real test area (UMBC CARDS outdoor testing site ~\cite{cards}). In this environment, we placed a static car obstacle on a pathway (Fig.~\ref{fig:virtual_physical_navigation}). The task for the robot is to navigate from a start point to a goal location while avoiding this obstacle and crossing a narrow wooden bridge, assuming the path is clear of other hazards. To challenge the virtual-physical coordination, we intentionally introduced some discrepancies between the virtual and real environments. For example, in the virtual world we added several fallen tree logs blocking the bridge, whereas in the actual environment the bridge is unobstructed. During the experiment, the physical robot and its virtual counterpart operate concurrently: the physical robot continuously receives navigation hints and environment updates from the virtual twin via the SERN Bridge, but it also cross-checks all virtual information against its own onboard sensors. In practice, the robot uses a real-time semantic segmentation module to detect obstacles in its camera/LiDAR data. If the virtual twin predicts an obstacle that the physical robot’s sensors do not actually see (or vice versa), the discrepancy is flagged and the robot adjusts its plan accordingly, always favoring real sensor data for safety. This way, the simulation can suggest alternate routes or warn of potential dangers, but the physical robot will not be misled by any incorrect or outdated virtual information. 

We compare SERN against a standard ROS setup that forwards the same ROS topics without SERN's bandwidth aware scheduling or predictor correction synchronization. For communication load tests, each active stream publishes high bandwidth point cloud data to stress the network. For synchronization and navigation, we report pose alignment error, communication latency, processing load, success rate, time to goal, and intervention count.

\begin{figure*}[!htb]
\centering
\includegraphics[width=\linewidth]{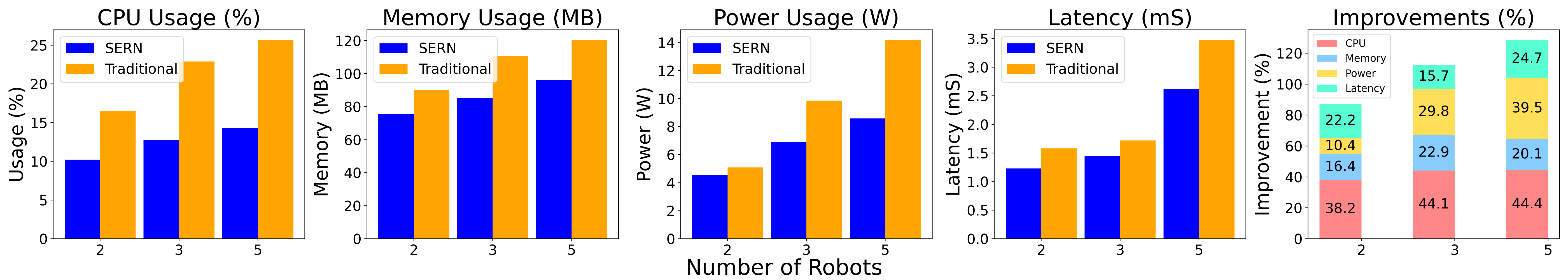}%
\caption{Scalability of the SERN bridge relative to a standard ROS setup under increasing replayed point cloud traffic (by increasing the robot numbers by 2, 3, and 5). The same stream configurations are applied to both systems, and the values are averaged over 20 runs for each load level. Lower latency and lower server-side resource usage indicate more efficient bridge behavior under the same communication load.}
\label{fig:aurora_comparison_agents}
\end{figure*}

\subsection{Synchronization Under Constrained Communication}

We first evaluate how well SERN keeps the virtual twin aligned with the physical robot when updates are delayed, dropped, or temporarily unavailable. The synchronization loop combines short horizon state propagation with correction from incoming telemetry. This is the part of the system that determines whether the virtual side remains useful when communication degrades. Under moderate delay and packet loss, the physical and virtual trajectories remained closely aligned throughout the run. The average steady state position error stayed below 5~cm, and the orientation error stayed below $2^\circ$. We also tested a 30~s communication blackout while the robot was moving. During the blackout, the virtual robot continued to advance using the last available control inputs. When communication resumed, the position error between the physical and virtual states was about 5~cm. In the same setting, a correction only baseline with a fixed gain PD controller and no predictive propagation drifted beyond 20~cm. Once updates resumed, SERN smoothly brought the virtual state back into alignment.

\begin{table}[!htb]
\centering
\scriptsize
\setlength{\tabcolsep}{8pt}
\begin{tabular}{p{2.0cm} p{2.0cm} c c}
\toprule
\textbf{Setting} & \textbf{Method} & \textbf{Pos. Err.} & \textbf{Rot. Err.} \\
\midrule
Moderate delay/loss & SERN & $<0.05$ m & $<2^\circ$ \\
30\,s blackout & Fixed gain PD & $>0.20$ m &$>3^\circ$ \\
30\,s blackout & SERN & $\approx 0.05$ m & $<3^\circ$ \\
\bottomrule
\end{tabular}
\caption{Synchronization summary under moderate delay, packet loss, and a 30\,s communication gap. The gap rows show the mismatch measured when communication resumed.}
\label{tab:sync_summary}
\end{table}

The system also maintained a consistent exchange between real and virtual environment updates. When a new obstacle was added in the virtual scene, the corresponding real object appeared at the expected location with an average offset of 2.3~cm. However, when the robot observed that the real bridge was clear while the virtual scene still showed fallen logs, that discrepancy was relayed back to the twin and used to correct the virtual state. In our setup, SERN pushed sensor driven updates into Unity at 10~Hz, which was sufficient for near real time alignment in the tested navigation task.

\subsection{Bridge Efficiency Under Increasing Load}

We next evaluate how the bridge behaves as the communication load increases. The goal of this experiment is not to claim a large scale multi robot field deployment. Instead, it tests whether bandwidth aware topic handling reduces unnecessary overhead while preserving the information needed to keep the twin usable. To do this, we replay recorded point cloud streams under separate namespaces and gradually increase the number of active streams. We then compare the SERN bridge and a standard ROS setup using end to end latency together with server side CPU, memory, and power usage. Figure~\ref{fig:aurora_comparison_agents} summarizes the results. Across the tested load levels, SERN reduces end to end latency by 15\% to 25\% and lowers processing load by about 15\% relative to the standard ROS baseline. CPU and memory usage also increase more gradually as the number of active streams rises. These results are consistent with the design of the bridge: by giving priority to synchronization critical messages and avoiding unnecessary forwarding, the bridge uses network and compute resources more efficiently under the same workload.

\begin{table}[t]
\centering
\scriptsize
\begin{tabular}{lccc}
\toprule
Mode & Success (\%) & Time to Goal (s) & Interventions (\#) \\
\midrule
Real Only & $85.0\pm7.5$\, (80.9--89.1) & $124.2\pm18.6$ & $1.3\pm0.6$ \\
Sim Only  & $70.0\pm10.3$\, (63.5--76.5) & $109.7\pm15.2$ & $1.9\pm0.8$ \\
SERN      & $\mathbf{95.0\pm5.0}$\, (92.2--97.8) & $\mathbf{98.5\pm12.4}$ & $\mathbf{0.4\pm0.5}$ \\
\bottomrule
\end{tabular}
\caption{Real Only, Sim Only, and SERN on a fixed navigation task with $n=20$ trials per mode. Values are reported as mean $\pm$ standard deviation, with 95\% confidence intervals in parentheses for success rate.}
\label{tab:baselines}
\end{table}

\begin{table}[t]
\centering
\scriptsize
\begin{tabular}{lcc}
\toprule
Policy & Teleoperation & Mapping \\
\midrule
Fixed Throughput  & 0.43 & 0.38 \\
Fixed Reliability & 0.56 & 0.63 \\
MMCF              & 0.39 & 0.32 \\
\bottomrule
\end{tabular}
\caption{Operational cost $J$ for each policy under latency sensitive teleoperation and bandwidth sensitive mapping profiles. Lower is better.}
\label{tab:mmcf_eval}
\end{table}

\begin{table*}[t]
\centering
\scriptsize
\begin{tabular}{lcccccc}
\toprule
Profile & Policy & Latency & Reliability & Compute & Bandwidth & Total Cost \\
\midrule
Teleoperation & Fixed Throughput  & 0.18 & 0.78 & 0.44 & 0.68 & 0.43 \\
Teleoperation & Fixed Reliability & 0.34 & 0.95 & 0.58 & 0.82 & 0.56 \\
Teleoperation & MMCF              & 0.21 & 0.91 & 0.41 & 0.52 & 0.39 \\
\midrule
Mapping & Fixed Throughput  & 0.24 & 0.80 & 0.36 & 0.49 & 0.38 \\
Mapping & Fixed Reliability & 0.42 & 0.97 & 0.63 & 0.88 & 0.63 \\
Mapping & MMCF              & 0.27 & 0.93 & 0.33 & 0.36 & 0.32 \\
\bottomrule
\end{tabular}
\caption{MMCF breakdown across two mission profiles. Each entry reports the normalized latency, reliability, compute, and bandwidth terms used in the total cost. Lower latency, compute, bandwidth, and total cost are better, while higher reliability is better.}
\label{tab:mmcf_breakdown}
\end{table*}

\subsection{Navigation Comparison: Real Only, Sim Only, and SERN}
\label{sec:baselines}

To measure the practical value of simulation enhanced navigation, we compare three operating modes on the same task: \textit{Real Only}, where the robot relies only on onboard sensing; \textit{Sim Only}, where the robot follows the virtual world model without live correction; and \textit{SERN}, where the robot uses the virtual twin for look ahead support while resolving disagreements with live sensor data. Each mode is evaluated over $n=20$ trials. We report the success rate, the time required to reach the goal, and the number of operator interventions, such as manual stops or resets when the robot becomes stuck. Table~\ref{tab:baselines} summarizes the results.

In the Sim Only setting, the robot sometimes makes poor decisions because the virtual world does not fully match the real scene. In our test case, the virtual logs near the bridge can prevent progress even though the real bridge is clear. Real Only avoids those false assumptions, but it has no preview of the upcoming layout and often progresses more cautiously, which increases time to goal. SERN combines the useful parts of both. The virtual twin provides early awareness of the parked car and the broader terrain layout, while live sensing prevents the robot from following stale or incorrect assumptions in the twin. This leads to the highest success rate, fewer interventions, and the shortest average time to goal. A video demonstration is available online.\footnote{Video demonstration available at: \url{https://pralgomathic.github.io/sern.multi-agent/}}

\subsection{MMCF Evaluation}

We evaluate the multi metric cost function used to choose bridge behavior at runtime. The cost function combines latency, reliability, compute load, and bandwidth usage,
\[
\begin{aligned}
J &= w_\ell\,\mathrm{latency} + w_r\,(1-\mathrm{reliability}) \\
  &\quad + w_c\,\mathrm{CPU} + w_b\,\mathrm{bandwidth}.
\end{aligned}
\]
where the weights depend on the mission profile. We compare MMCF against two fixed bridge policies: \textit{Fixed Throughput}, which favors high compression and low redundancy, and \textit{Fixed Reliability}, which favors lower compression and higher redundancy. We evaluate two representative profiles. The \textit{Teleoperation} profile places more weight on low latency, while the \textit{Mapping} profile places more weight on bandwidth efficiency. Table~\ref{tab:mmcf_eval} summarizes the overall operational cost under the teleoperation and mapping profiles. In both cases, MMCF gives the lowest cost, indicating that the selected runtime policy provides a better balance across competing communication and compute objectives than either fixed baseline. To make this behavior easier to interpret, Table~\ref{tab:mmcf_breakdown} shows the normalized latency, reliability, compute, and bandwidth terms that contribute to the final cost. We define reliability as the fraction of synchronization critical messages that are delivered within a timeline $\delta$:
\[
R = \frac{N_{\mathrm{on\text{-}time}}}{N_{\mathrm{sent}}}.
\]
Higher $R$ is better, so the cost uses the term $(1-R)$.

\subsection{Runtime Reconfiguration Evaluation}

\begin{figure}[t]
    \centering
    \includegraphics[width=\linewidth]{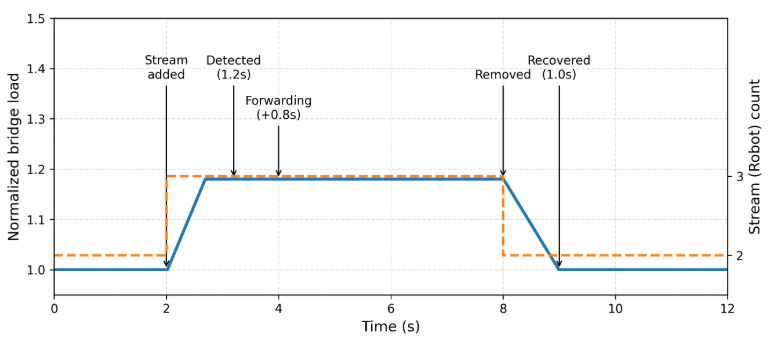}
    \caption{Runtime reconfiguration demonstration. The dotted curve shows the number of active robot streams, and the solid curve shows normalized bridge load as a new stream is added and later removed.}
    \label{fig:runtime_reconfig_demo}
\end{figure}

We also evaluated whether the bridge can incorporate newly appearing streams at runtime without manual restart or static reconfiguration. In this demonstration, the system started with two active robot streams on one subnet and a set of monitoring nodes on the other, with dynamic topic discovery enabled. An additional stream was then introduced at runtime. The bridge detected the new publisher topics automatically and began forwarding them through the AuroraXR server without interrupting the already active streams. Figure~\ref{fig:runtime_reconfig_demo} summarizes this behavior. The dotted curve shows the number of active robot streams, while the solid curve shows the normalized bridge load. The system began with two active streams. At \(t=2\) s, a new stream was introduced, increasing the active stream count from two to three. The bridge detected the new stream within approximately 1.2\,s and began forwarding its topics about 0.8\,s later. During this transition, the load on the bridge increased moderately but remained stable after forwarding began. At \(t=8\) s, the added stream was removed, and the bridge returned to its original load level within about 1.0\,s, again without manual restart or reconfiguration. This behavior is useful in practice because it allows the communication layer to adapt to changing network composition while preserving the same synchronization and routing logic.

\section{Conclusion, Limitations, and Future Directions}

We presented \textbf{SERN}, a cross reality framework that connects a high fidelity virtual twin with a physical robot through a bandwidth adaptive bridge and a physics aware synchronization pipeline. Across communication constrained settings, SERN reduced message latency and processing load while keeping the virtual state closely aligned with the physical system. The results also showed that combining predictive synchronization with live correction can improve navigation robustness compared with using only the real or virtual side alone. However, this work has several limitations. The finite interval error bound applies during a communication gap and depends on the control mismatch remaining bounded over that interval. Under sharp maneuvers or longer disruptions, the mismatch can grow and the bound becomes less informative until fresh updates arrive. In addition, our field validation used one physical robot, while added traffic at higher load was generated through replayed ROS bag streams. Future work will extend the system to larger physical robot teams, test it across a wider range of environments and link conditions, and improve how the virtual twin selects critical updates and scene detail online. Another useful direction is to reduce runtime compute and energy cost while maintaining the same level of synchronization quality.


\bibliographystyle{unsrt}
\bibliography{bibliography}

\end{document}